\documentclass[runningheads]{llncs}

\usepackage{caption}
\usepackage{subcaption}
\usepackage[T1]{fontenc}
\usepackage{bbding}
\usepackage{cite}
\usepackage{amsmath,amssymb,amsfonts}
\usepackage{algorithmic}
\usepackage{graphicx}
\usepackage{textcomp}
\usepackage{xcolor}
\usepackage{amsmath}
\usepackage{pifont}

\usepackage[utf8]{inputenc} %
\usepackage[T1]{fontenc}    %
\usepackage{hyperref}       %
\usepackage{url}            %
\usepackage{booktabs}       %
\usepackage{amsfonts}       %
\usepackage{nicefrac}       %
\usepackage{microtype}      %
\usepackage{algorithm}
\usepackage{bm}
\usepackage{multirow}

\usepackage{wrapfig}
\usepackage{lipsum} 
\begin{document}
\title{Balanced and Explainable Social Media Analysis for Public Health with Large Language Models}
\titlerunning{ALEX}
\author{Yan Jiang \and
Ruihong Qiu\and
Yi Zhang\and
Peng-Fei Zhang}
\institute{The University of Queensland\\
\email\{yan.jiang, r.qiu, y.zhang4, pengfei.zhang\}@uq.edu.au}
\authorrunning{Jiang et al.}
\maketitle             

\begin{abstract}
As social media becomes increasingly popular, more and more public health activities emerge, which is worth noting for pandemic monitoring and government decision-making. Current techniques for public health analysis involve popular models such as BERT and large language models (LLMs). Although recent progress in LLMs has shown a strong ability to comprehend knowledge by being fine-tuned on specific domain datasets, the costs of training an in-domain LLM for every specific public health task are especially expensive. Furthermore, such kinds of in-domain datasets from social media are generally highly imbalanced, which will hinder the efficiency of LLMs tuning. To tackle these challenges, the data imbalance issue can be overcome by sophisticated data augmentation methods for social media datasets. In addition, the ability of the LLMs can be effectively utilised by prompting the model properly. In light of the above discussion, in this paper, a novel ALEX framework is proposed for social media analysis on public health. Specifically, an augmentation pipeline is developed to resolve the data imbalance issue. Furthermore, an LLMs explanation mechanism is proposed by prompting an LLM with the predicted results from BERT models. Extensive experiments conducted on three tasks at the Social Media Mining for Health 2023 (SMM4H) competition with the first ranking in two tasks demonstrate the superior performance of the proposed ALEX method. Our code has been released in \url{https://github.com/YanJiangJerry/ALEX}.

\end{abstract}

\section{Introduction}
Social media mining with natural language processing (NLP) techniques has great potential in analysing public health information~\cite{al2022role}. For example, mining social media posts about vaccines can indicate the vaccine acceptance rate, and analysing public opinions towards healthcare can influence government policies~\cite{COVID}. 

Despite that there are abundant health-related datasets on social media, their value in assisting public health analysis has been largely untapped~\cite{inproceedings}. Current techniques for public health analysis generally involve various transformer-based models, such as BERT~\cite{bert} and LLMs~\cite{chinchilla,instructgpt, llama, glm, u2l, openai2023gpt4, DBLP:journals/corr/abs-2303-18223}. With a strong ability for text processing, such models can effectively extract valuable information from social media for various kinds of tasks such as text classification.

Although existing techniques have shown remarkable performance on some of the tasks, they often encounter problems when classifying public health information on social media. Firstly, the extremely large parameter sizes and high computational requirements for adopting LLMs in specific public health areas will increase their training difficulty, while smaller models such as BERT usually have limited input token size which may also make training difficult to include all the information. Secondly, imbalanced data also poses a significant challenge for current techniques, which may lead to biased predictions~\cite{kumar2021classification}. Lastly, in the social media datasets, it is common to witness a significant variation in the importance of different categories. For instance, when analysing social anxiety diagnosis, the positively diagnosed cases would hold greater analytical value than the others. Existing systems do not fully consider the above problems, therefore, the performance of public health analysis will be seriously limited.

In order to address the aforementioned problems, an effective method for public health text classification is urgently needed. In this paper, firstly, a complete data augmentation pipeline and a weighted loss fine-tuning are employed to address the imbalance problem in social media. Secondly, to reduce the computational cost of LLMs training and solve BERT's limited token size problem, an explanation and correction mechanism based on LLMs inference is introduced. Specifically, the predicted labels from BERT and the original text are prompted together for an LLM to examine by taking the given predicted labels to find the existence of evidence in the text thus explaining whether the given labels are correct or not. Such an innovative approach successfully incorporates the language generation capabilities of LLMs into text classification tasks, thus enhancing the credibility and performance of the results. This framework is named by \textbf{A}ugmentation and \textbf{L}arge language model methods for \textbf{EX}plainable (ALEX) social media analysis on public health. The main contributions can be summarised as follows:

\begin{itemize}
\item A balanced training pipeline is proposed by data augmentation and weighted-loss fine-tuning to address the class and importance imbalance problems.

\item An LLMs-based method is proposed to conduct post-evaluation on the predicted results, which successfully leverages LLMs' ability without training.

\item The ALEX model achieves number one in two shared tasks (Task 2 and Task 4) in the SMM4H 2023 competition and a top result in Task 1~\cite{jiang2023uq} which are all text classification tasks related to public health on social media datasets.
\end{itemize}

\section{Related Work} \label{sec:2}
\subsection{Text Classification}
Text classification refers to classifying given sequences into predefined groups. Prior studies have highlighted the performance of recurrent models ~\cite{liu2016recurrent,seo2018neural,yogatama2017generative}and attention mechanisms~\cite{vaswani2017attention} for developing text representation~\cite{liu2005text}. Building upon those ideas, extensive research has substantiated the effectiveness of employing pre-trained models~\cite{qiu2020pre} such as BERT~\cite{devlin2019bert} and OpenAI GPT~\cite{radford2018improving,radford_language_2019,brown2020language,openai2023gpt4} on text classification. The BERT model is a multi-layer bidirectional Transformer~\cite{bert} that has been proven to achieve state-of-the-art performance across a wide range of NLP applications~\cite{sun2020finetune}. After the advent of BERT, RoBERTa~\cite{liu2019roberta} is proposed to resolve BERT's undertrained problem. ALBERT is invented to reduce the computational resources of BERT~\cite{lan2020albert}. DistilBERT, a distilled version of BERT, showed that it is possible to reduce the size of a BERT model by 40 percent~\cite{sanh2020distilbert}. The combination of BERT with other architectures such as BERT-CNN~\cite{safaya2020kuisail}, BERT-RCNN~\cite{bertRCNN}, and  BERT-BiLSTM~\cite{9515273} are also been proved to be capable to improve the classification accuracy for social media analysis. 

In addition to structural improvement, BERT can also be fine-tuned on different domain-specific datasets. BERTweet is the first open-source pre-trained language model designed for English Tweets~\cite{nguyen2020bertweet}. Similarly, TwHIN-BERT ~\cite{zhang2022twhinbert}, Camembert~\cite{DBLP:conf/acl/MartinMSDRCSS20} and Flaubert~\cite{DBLP:conf/lrec/LeVFSCLACBS20} have also achieved significant performance improvements for text classification in specific domains.

\subsection{Large Language Models}
LLMs are pre-trained on large corpus and exhibit a strong ability to generate coherent text. GPT-1~\cite{radford_language_2019} is the first LLM to follow the transformer architecture~\cite{vaswani2017attention} based on the decoder in 2017. Based on it,  GPT-2 enlarges the parameter size to 1.5B and begins to learn downstream tasks without explicit training~\cite{radford_language_2019}. GPT-3 continues to scale the number of parameters to 175B with few shot learning~\cite{brown2020language}. Such strong models can be applied for inference without any gradient updates. Meanwhile, GPT-3.5 (ChatGPT) is invented to train on up-to-date datasets. In March 2023, OpenAI released the development of GPT-4~\cite{openai2023gpt4}, a multimodal LLM capable of accepting both image and text inputs. Researchers have suggested the potential utility of LLMs in health fields~\cite{singhal2022large}. For example, in disease surveillance tasks, such models can be employed by analysing large volumes of social media data to provide insights for detecting and monitoring disease outbreaks~\cite{ZENG2021437}. 

\section{Preliminaries: Task Definition}
For all three tasks, the problem can be framed as text classifications with binary classification for Task 1 and 4 and three-class classification for Task 2. For each task, given a set of social media post texts $\{x_1,...,x_n\}\subseteq \mathcal{X}$ and the corresponding labels $\{y_1,...,y_n\}\subseteq \mathcal{Y}$, the objective is to develop a model that can effectively classify the input social media texts. For all the following equations, bold lowercase letters denote vectors, lowercase letters denote scalars or strings and uppercase letters denote all the other functions. 

\section{Methodology}
\begin{figure}[!t]
  \centering
  \includegraphics[width=0.8\textwidth]{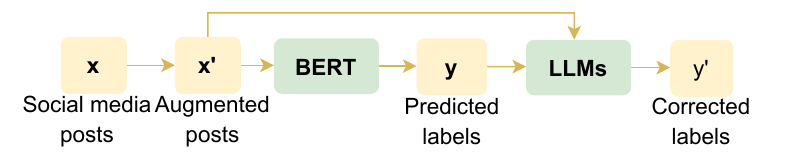}
  \caption{Overall framework of ALEX method. Firstly, the social media posts are augmented to fine-tune the BERT models. The labels are then predicted by BERT. After combining the labels with their original posts and constructing a complete prompt by adding instructions and examples, LLMs can take the prompt to explain and identify those incorrectly predicted labels from BERT.}
  \label{fig1}
\end{figure}

In this section, the ALEX method is divided into several components as Figure \ref{fig1} shows. Firstly, balanced training involves augmentation is employed to solve the imbalance problem. Secondly, fine-tuning with the weighted loss is conducted on BERT. Lastly, an LLM is applied to explain and correct the labels. 

\subsection{Augmentation and Resampling}
For text classification in social media, data imbalance problems may significantly influence the models' performance. The TextAttack~\cite{morris2020textattack}, which is a well-developed framework designed for adversarial attacks, data augmentation, and adversarial training, is chosen to enrich the number of the minority class because it can automatically incorporate adversarial attacks to enhance the robustness of data augmentation with just a few lines of code. After data augmentation, oversampling and undersampling are also introduced to make the number of each class exactly the same as Equation \ref{eq1} shown:
\begin{equation} \label{eq1}
\textbf{x}'= \text{Resample}(\text{Aug}(\textbf{x})),
\end{equation}
where $\textbf{x}'\in\mathbb{R}^{n}$ denotes the augmented text. Aug is the TextAttack augmentation and Resample denotes oversampling and undersampling methods.

\subsection{Social Media Text Encoding}
After getting the balanced dataset, those texts will be encoded by BERT to get better embeddings for text classification as Equation \ref{eq2} shows. The BERT's output sequence can be divided into segments with the first token being [CLS], which is the special classification token for predicting the labels' probability.
\begin{equation} \label{eq2}
\textbf{e} = \text{BERT}(\textbf{x}'),
\end{equation}
where $\textbf{e}\in\mathbb{R}^{d}$ represent the embedding of the [CLS] token while $d$ represents its dimension. To employ the BERT's embedding for text classification, a straightforward approach is adding an additional classifier as Equation \ref{eq3} shows:
\begin{equation} \label{eq3}
\hat{y} = \text{argmax}(\text{Classifier}(\textbf{e})),
\end{equation}
where $\hat{y}\in\mathbb{R}$ is the label predicted by BERT. The output from the classifier represents the probability of each label, then the argmax function will extract the index that has the max probability, which will be the predicted label.

\subsection{Weighted Loss}

In social media analysis for public health, there is usually a tendency to place importance on specific classes in text classification. For example, while analysing the number of COVID cases, the F1 score for the positive class holds greater analytical values compared to the negative class. To address this issue, the weighted binary cross-entropy loss will be adopted, which allows for the loss weight adjustments on different labels based on their importance. Such importance can be manually decided based on the task objectives. In this way, we can ensure the model assigns more emphasis to a specific label during the training process. Based on the importance, the loss weight $\lambda$ can be introduced, which will be integrated into binary cross-entropy loss~\cite{loss} as Equation \ref{eq5} shows:
\begin{equation} \label{eq5}
\ell = -\frac{1}{n} \sum_{i=1}^{n} \lambda  y  \text{log}(\hat{y}) + (1 - y) \text{log}(1-\hat{y}).
\end{equation}

\begin{wrapfigure}{r}{0.6\textwidth}
  \centering \vspace{-20pt}
  \includegraphics[width=0.6\textwidth]{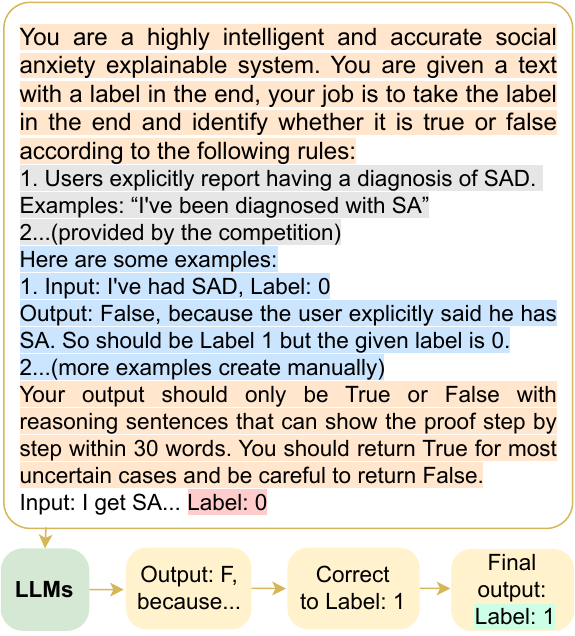}
  \caption{LLM explanation and correction. The prompt includes instructions (orange), labelling rules (grey), examples (blue), and the input text concatenated with its wrong label predicted by BERT in red. The LLMs successfully identified the label as False therefore the label will be corrected to the final correct label in green.}
  \label{ALEX}
  \vspace{-30pt}
\end{wrapfigure}

\subsection{LLM Explanation and Corrections at Inference Time}\label{llm}
To overcome the limited input size of BERT and enhance its performance, LLMs will be introduced to further correct the prediction using their powerful language modelling ability. Firstly, the predicted label from BERT will be concatenated with the original text. Secondly, the processed text will be combined with the labelling rules and manually created instructions for each task with a few representative examples to form a complete prompt for LLMs' few-shot prompting as Figure \ref{ALEX} shows:

By constructing  BERT's prediction into a prompt, LLMs can seek logical proofs for the label within the original input. If LLMs successfully find supporting evidence in the original text that aligns with the given label, it will return ``True'' with a concise provable explanation that can explain the reason. However, \\if LLMs fail to identify any compelling evidence that could support the given label or find the opposite proofs that can overturn the given label, it will return ``False'' and will use the Chain of Thought (CoT)~\cite{chain} idea to give a step-by-step explanation for why it considers the label is incorrect as Equation \ref{eq6} shows:
\begin{equation} \label{eq6}
\text{exp} = \text{LLMs}(\text{Prompt}(x,\hat{y})),
\end{equation}
where exp refers to the explanation including the judgments of whether LLMs think the label is true with the step-by-step proof. Prompt represents the prompt construction including the concatenation of the original text with the BERT predicted label, adding labelling rules with manually created instructions, and a few examples.

Finally, to utilise the LLMs' explanation, a correction mechanism is designed. For cases when LLMs return ``True'' in the explanation, the original labels from BERT will remain. For cases when the LLM returns ``False'', there are two options. Firstly, if the number of ``False'' cases is relatively small, the manual review can be achievable to assess the LLMs' explanation. If the evidence is compelling, the label will be corrected to the others. Such a method will largely reduce the manual cost compared with the fully manual labelling method by leveraging the LLMs' language comprehension ability to screen out suspicious samples in advance. The second option is directly modifying the ``False'' label to the other label which will minimise the manual efforts. For binary text classification tasks, the ``False'' label can be directly converted to the other label. For the three-class text classification task, the ``False'' label can be converted to the majority label.

\section{Experiments}
\subsection{Experimental Setup}
\subsubsection{Datasets.}
This study employed three public health-related social media datasets provided by the SMM4H 2023 workshop. For \textbf{Task 1}, the dataset involves 600,000 English-language tweets excluding retweets by Twitter streaming API\footnote{\url{https://developer.twitter.com/en/docs/twitter-api}}. Such tweets are selected based on the presence of keywords indicating COVID-19 testing, diagnosis, or hospitalisation~\cite{data1}. Similarly, the dataset for \textbf{Task 2} is also a Twitter dataset that is labelled by three sentiments associated with therapies: negative, neutral, and positive. Lastly, the dataset for \textbf{Task 4} is extracted from the Reddit platform\footnote{\url{https://www.reddit.com/}} and is divided into Label 0 for negative social anxiety cases and Label 1 for positive cases. The statistics of SMM4H competition are shown in Table \ref{tab2}. In this paper, only the validation set is evaluated since the competition only provides labels for the validation set instead of the testing set.

\begin{table}[!t]
\centering
\caption{Statistics of the public health datasets from SMM4H 2023.}\label{tab2}
\resizebox{0.8\linewidth}{!}{
\begin{tabular}{c|c|c|c|c|c|c|c}
\toprule
Dataset &  Classes & Type & Average & Max & Train & Validate & Test \\
\midrule
Task 1 &  2 & COVID Diagnosis &38.1&106&7600&400&10000\\
\midrule
Task 2 &  3 & Sentiment Analysis &32.7&100&3009&753&4887\\
\midrule
Task 4 & 2 & Social Anxiety Analysis&235.2&2000&6090&680&1347\\
\midrule
\end{tabular}
}
\end{table}

\subsubsection{Metrics.}
Popular evaluation metrics such as precision, recall, and F1 score are used. For Task 1 and Task 4, only the F1 score for Label 1 is evaluated. For Task 2, the micro-averaged F1 score is evaluated as required by the SMM4H competition which is the same as the accuracy. Therefore, in all the experiments, the F1 score for Label 1 and the accuracy are observed. For the baseline experiment, the macro average F1 score is also evaluated to show the imbalanced problem effect.

\subsubsection{Implementation.}
Firstly, the TextAttack will be implemented to augment the number of minority class samples. Then, the training set will be resampled which will make the number of each label balanced. For the fine-tuning, the implementation is based on the Transformers open-source library~\cite{wolf2020huggingfaces}. For hyperparameters optimisation, the batch size from the range \{4, 8, 16\} is tested and the AdamW optimiser~\cite{loshchilov2019decoupled} is implemented with a learning rate choosing from 2e-5 to 1e-4. To mitigate overfitting problems, the weight decay from \{0.001, 0.005, 0.01\} is used. The number of warm-up steps is set to zero and the experiments are conducted with the number of epochs within the range of \{2, 4, 6, 8, 10, 12, 16, 20\}. The loss weight $\lambda$ is chosen from the range of \{1, 1.5, 2, 2.5, 5\}. For the baseline models, the batch size 8, learning rate 2e-5, weighted loss 0.1, and epoch 6 are used except for the XLNet model which uses batch size 4 due to the hardware limitations. For XLNet which has a larger token length, the inputs will not be truncated. However, for BERT-related models, the input will be head truncated~\cite{sun2020finetune} to be within 512 tokens for the large model and 128 tokens for the base model.

In Task 4, BERTweet-Large experiments indicate that a learning rate of 2e-5, a batch size of 16, a weight decay of 0.005, and epochs of 6 will lead to the highest F1 scores for Label 1. For other tasks, weight decay of 0.01 and epoch 3 will be adopted as over-training may hinder the models' performance. For the LLMs part, GPT-3.5 is employed from OpenAI\footnote{\url{https://openai.com/}} to explain and correct the BERT's prediction as discussed in Section \ref{llm}. For Task 1 and Task 4, the ``False'' Label 1 sample will be automatically converted to Label 0. While for Task 2, the ``False'' Label 0 predictions will be converted to the majority class Label 1.

\subsubsection{Baselines.} 
The following baselines are chosen for comparison:
\begin{itemize}
\item[$\bullet$] \textbf{BERT}~\cite{bert} is the original baseline for all text classification Tasks which is pre-trained on the general corpus. Both base and large models are tested.

\item[$\bullet$] \textbf{RoBERTa}~\cite{liu2019roberta} is a compact and efficient version of the BERT, and it is designed for fast and resource-friendly NLP Tasks.

\item[$\bullet$] \textbf{XLNet}~\cite{DBLP:journals/corr/abs-1906-08237} is a highly effective language model that leverages the Transformer-XL architecture with both autoregressive and permutation-based training. It allows for larger token size input than BERT.

\item[$\bullet$] \textbf{BERTweet}~\cite{nguyen2020bertweet} is a baseline for text classification Tasks on social media datasets like Twitter. Both the BERTweet-base and BERTweet-large models are experimented with.

\item[$\bullet$] \textbf{CT-BERT (v2)}~\cite{müller2020covidtwitterbert}\footnote{\url{https://huggingface.co/digitalepidemiologylab/covid-twitter-bert-v2}} is a BERT-Large model and it has been fine-tuned on more COVID Twitter datasets than CT-BERT (v1). As CT-BERT does not involve the base model, only the large model is evaluated.
\end{itemize}

In overall comparison, the ``-B'' indicates the base models with the ``-L'' denotes the large models. In all following experiments, only BERTweet-Large is used for ALEX-Large in Task 2 and Task 4 as well as CT-BERT (v2) is used for Task 1.

\begin{table}[!t]
\centering
\caption{Overall model performance.}\label{tab3}
\resizebox{0.9\linewidth}{!}{
\begin{tabular}{c|ccc|ccc|ccc}
\toprule
\multirow{2}{*}{Model} & \multicolumn{3}{c|}{Task 1} & \multicolumn{3}{c|}{Task 2}& \multicolumn{3}{c}{Task 4}\\
\cline{2-10}
& F1  &  Accuracy & M-Avg  &  F1  & Accuracy & M-Avg & F1 & Accuracy & M-Avg \\
\midrule
BERT-B & 82.25 & 93.15 & 93.74 & 83.33 & 71.24 & 58.94 & 79.75 & 84.15 & 83.37\\
\midrule
BERT-L & 85.75 & 95.02 & 93.79 & 80.44 & 69.30 & 64.92 & 81.17 & 85.74 & 84.85\\
\midrule
RoBERTa-B & 91.84 & 92.77 & 91.01 & 77.33 & 70.07 & 66.45 & 81.86 & 87.13 & 85.87\\
\midrule
RoBERTa-L & 92.71 & 94.97 & 93.56 & 81.35 & 74.45 & 71.07 & 44.52 & 72.74 & 63.22\\
\midrule
XLNet-B & 3.04 & 64.56 & 51.47 & 84.82 & 76.28 & 76.87 & 78.92 & 82.73 & 82.14 \\
\midrule
XLNet-L & 0.0 & 64.33 & 50.37 & 85.00 & 75.74 & 76.78 & 84.89 & \underline{89.18} & \underline{87.66}\\
\midrule
BERTweet-B & 88.31 & 93.67 & 92.82 & 79.97 & 72.44 & 68.37 & 80.77 & 83.49 & 82.93\\
\midrule
BERTweet-L & 91.22 & 95.57 & 93.61 & \underline{85.61} & 73.67 & 77.49 & \underline{85.42} & 87.01 & 85.38\\
\midrule
CT-BERT (v2) & \underline{93.11} & \underline{95.62} & \underline{94.47} & 85.02 & \underline{76.43} & \underline{77.81} & 61.93 & 70.29 & 68.78 \\
\midrule
\midrule 
ALEX-B (ours) & 90.77 & 92.04 & 91.25 & 82.20 & 73.00 & 71.19 & 83.39 & 88.25 & 87.15\\
\midrule
ALEX-L (ours) & \textbf{94.97} & \textbf{96.71} & \textbf{95.84} & \textbf{89.13} & \textbf{77.84} & \textbf{79.57} & \textbf{88.17} & \textbf{89.84} & \textbf{88.26}\\
\bottomrule
\end{tabular}
}
\end{table}
\subsection{Overall Performance}
The overall performance of baseline methods on the validation set is presented in Table \ref{tab3}. The first finding is that the original BERT models may outperform large models such as XLNet, which may owing to the serious impact of the imbalanced datasets in Task 1. Generally, for BERT-related models, the large model may outperform the base model. However, XLNet-Large underperforms XLNet-Base and gets a zero score for Label 1's F1 score in Task 1 which may related to the complexity of the model. Although XLNet models get generally low performance on Task 1 compared to other methods, their accuracy and macro-average F1 for Task 2 and Task 4 are remarkable. For RoBERTa, experiment results show that it can get better results compared to BERT on Task 1 and Task 4. However, for the base model, RoBERTa underperforms BERT in Task 2. 

Generally, in Task 1 and Task 4 where Label 1 is the minority class, the F1 score for Label 1 is observed to be lower than the overall macro F1 score, while for Task 2 where Label 1 is the majority class, the F1 score for Label 1 is observed to be higher than the overall macro F1 score. It is noticeable to see that BERTweet-Large model performances are generally better than others on the F1 score for Label 1 in Task 2 and Task 4. Moreover, for Task 1, CT-BERT (v2) gets the highest F1 score for Label 1 compared to all other baselines. Our ALEX-Large method outperforms all other methods in three tasks as it integrates balanced training and LLMs correction.

\begin{wraptable}{r}{5.cm}
\centering \vspace{-45pt}
\caption{Ablation study for the ALEX method.}\label{tab6}
\resizebox{\linewidth}{!}{
\begin{tabular}{c|c|cc|cc|cc}
\toprule
\multirow{2}{*}{Bal} & \multirow{2}{*}{LLMs} & \multicolumn{2}{c|}{Task 1} & \multicolumn{2}{c|}{Task 2}& \multicolumn{2}{c}{Task 4}\\
\cline{3-8}
& & F1 & Acc & F1 & Acc & F1 & Acc \\
\midrule
\XSolid & \XSolid & 93.11 & 95.62 & 85.61 & 73.67 & 85.42 & 87.01\\
\midrule
\Checkmark & \XSolid & \textbf{94.97} & \textbf{96.71} & 86.17 & 74.34 & 86.64 & 87.94\\
\midrule
\Checkmark & \Checkmark & 93.24 & 95.11 & \textbf{89.13} & \textbf{77.84} & \textbf{88.17} & \textbf{89.84}\\
\bottomrule
\end{tabular}
}\vspace{-20pt}
\end{wraptable}

\subsection{Ablation Study}
In order to validate the effectiveness of ALEX's method, an ablation study is conducted. As shown in Table \ref{tab6}, firstly, for all the tasks, the effectiveness of balanced training (``Bal'') is verified by observing improvements in the balanced training compared to baseline methods that use imbalanced training.

Another component of ALEX is the LLMs explanation and correction ``LLMs''. As GPT-3.5 is a language model and our objective is to leverage its language understanding ability to explain and correct the predictions, we do not test its performance directly on the text classification tasks. To simplify the process, the GPT-3.5 is constrained to only explain and correct Label 1's samples from BERT's prediction which aims to improve Label 1's precision. For Task 2, GPT-3.5 is constrained to only verify Label 0 predictions and correct them to the majority class which is Label 1. Experiment results reveal that GPT-3.5's explanation and correction can improve Task 2 and Task 4 Label 1's F1 scores from about 86 to above 88 compared to the results after balanced training. 

However, as for Task 1, because the F1 score for Label 1 after balanced training has already achieved a high score of about 95, GPT-3.5 struggles to further identify the few false positive predictions from BERT output, thus resulting in decreased performance compared to the results that use balanced training only. The possible reasons will be analysed from two aspects which are the prompts effect in Section \ref{prompt} and the LLMs hallucinations impact in Section \ref{hallucination}. Therefore, only the balanced training method from ALEX is used in Task 1.

\subsection{Strategies for Balanced Training}\label{loss}
In this study, the effects of different methods for balanced training are compared. In the table, ``Focal'' refers to fine-tuning BERT with focal loss while  ``Weighted'' means fine-tuning solely using weighted loss. ``Aug'' refers to data augmentation and ``Balanced'' refers to the combination of augmentation and weighted loss fine-tuning. The loss weight is set to 2 for Task 1 and Task 4. For Task 2, the weight for each class is set to be 2:1:2 according to experiment results. 

As shown in Table \ref{tab5}, it is noticeable to see that focal loss fine-tuning without augmentation will lead to decreased performance on all three tasks while weighted loss fine-tuning may increase the F1 score and accuracy for Task 1. However, the improvement for Task 2 and Task 4 is not obvious only with the weighted loss fine-tuning. After combining the augmentation with weighted loss fine-tuning, the balanced training method can produce superior performance in all three Tasks, leading to a visible improvement for Task 2 and Task 4. Visualisation for the effectiveness of the balanced training is shown in Section \ref{vis}.

\begin{table}[!t]
\centering
\caption{Strategies for balanced training.}\label{tab5}
\resizebox{0.6\linewidth}{!}{
\begin{tabular}{c|cc|cc|cc}
\toprule
\multirow{2}{*}{Balanced Methods} & \multicolumn{2}{c|}{Task 1} & \multicolumn{2}{c|}{Task 2}& \multicolumn{2}{c}{Task 4}\\
\cline{2-7}
& F1 & Acc & F1 & Acc & F1 & Acc \\
\midrule
Aug & 93.31 & 95.98 & 85.98 & 74.11 & 85.65 & 87.34\\
\midrule
Focal & 91.04 & 93.05 & 77.93 & 68.45 & 82.91 & 84.74\\
\midrule
Weighted & 94.81 & \textbf{97.14} & 85.39 & 73.10 & 85.47 & 86.07\\
\midrule
Balanced & \textbf{94.97} & 96.71 & \textbf{86.17} & \textbf{74.34}  & \textbf{86.64} & \textbf{87.94}\\
\bottomrule
\end{tabular}
} \vspace{-10pt}
\end{table}

\begin{wrapfigure}{r}{6.25cm}\vspace{-23pt}
  \begin{tabular}{cc}
    \begin{subfigure}[b]{0.25\textwidth}
      \includegraphics[width=\textwidth]{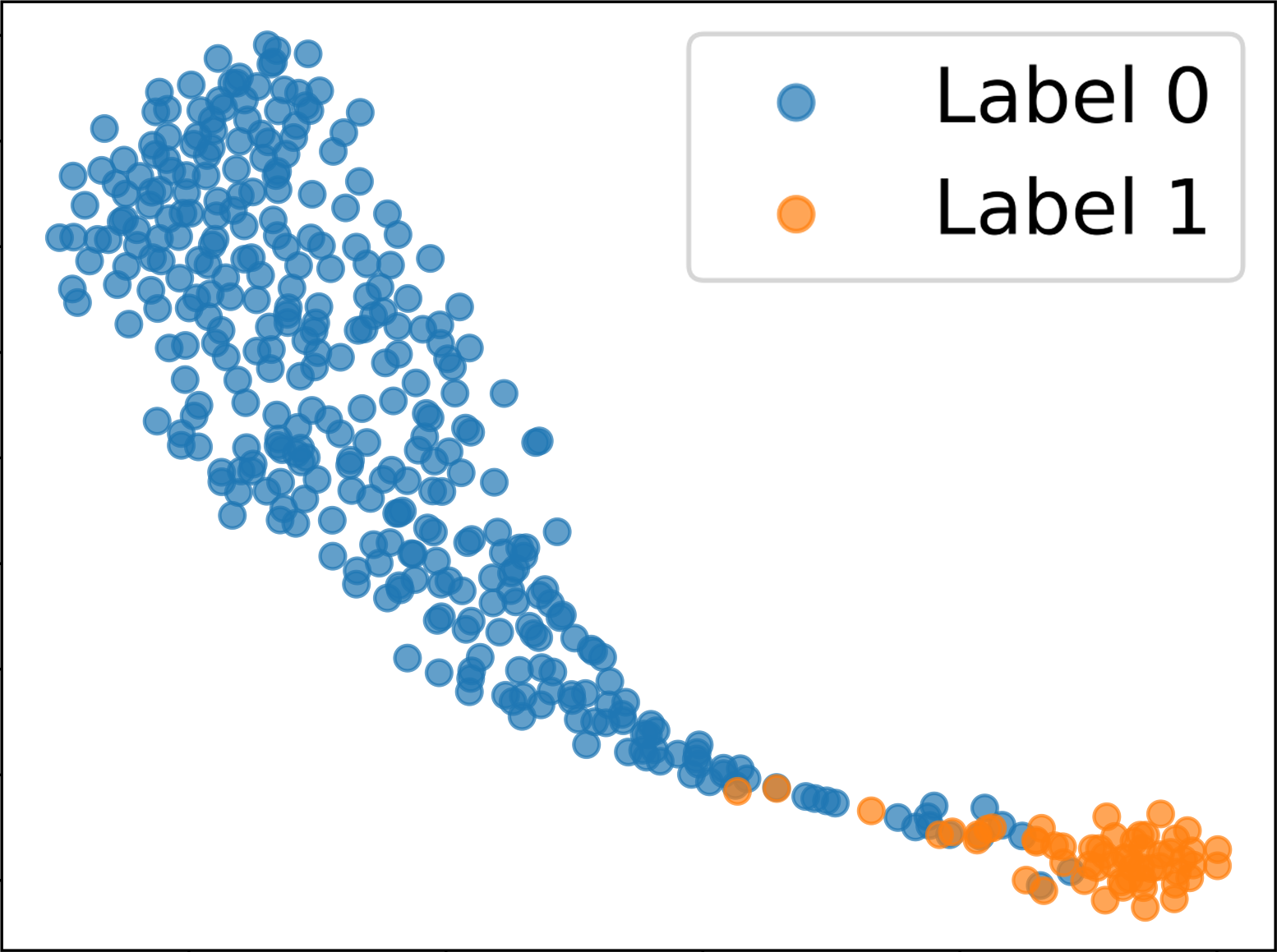}
      \caption{Task 1.}
    \end{subfigure}
    &
    \begin{subfigure}[b]{0.25\textwidth}
      \includegraphics[width=\textwidth]{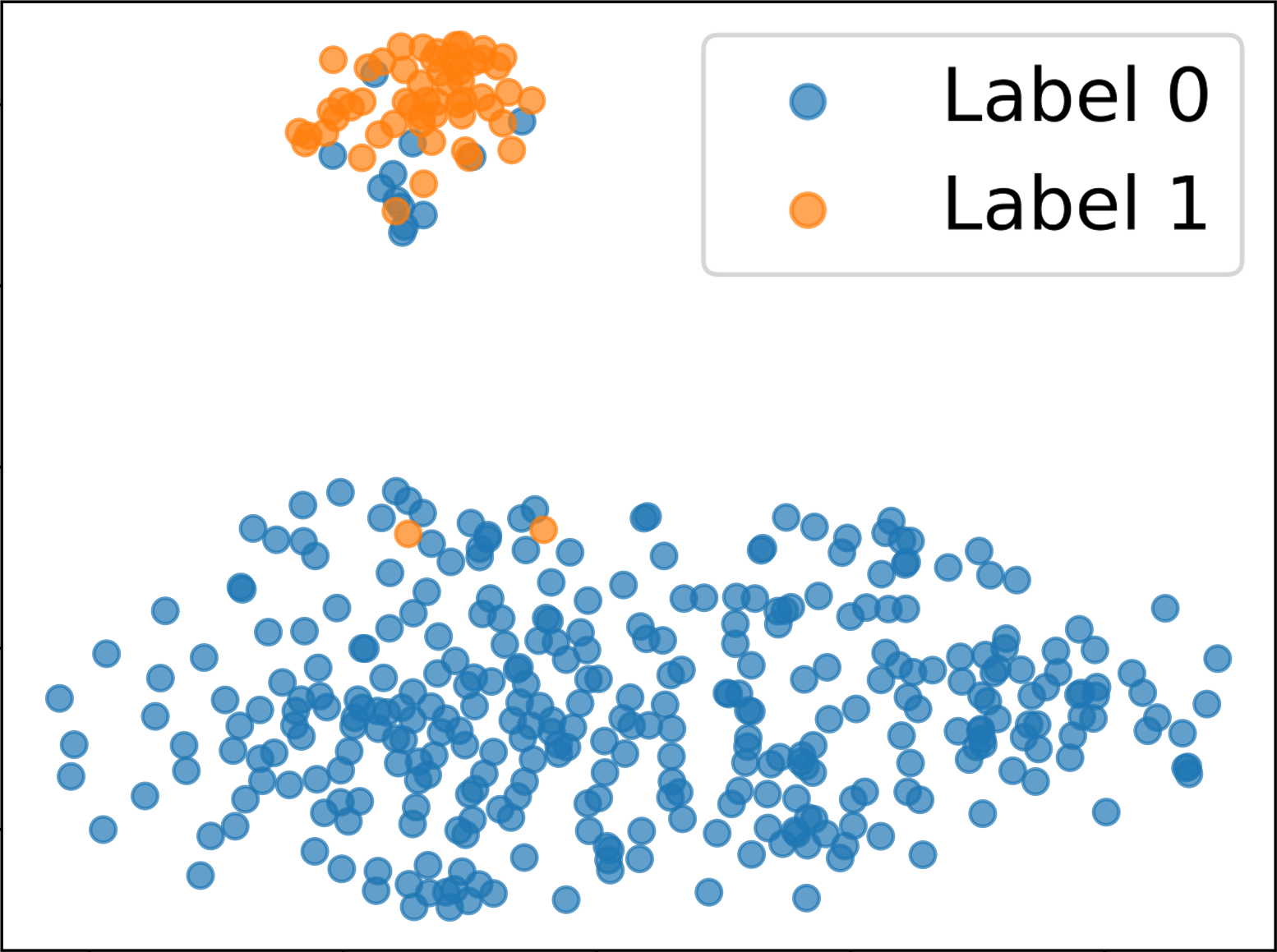}
      \caption{Task 1 (Balanced).}
    \end{subfigure}
  \end{tabular}
  
  \begin{tabular}{cc}
    \begin{subfigure}[b]{0.25\textwidth}
      \includegraphics[width=\textwidth]{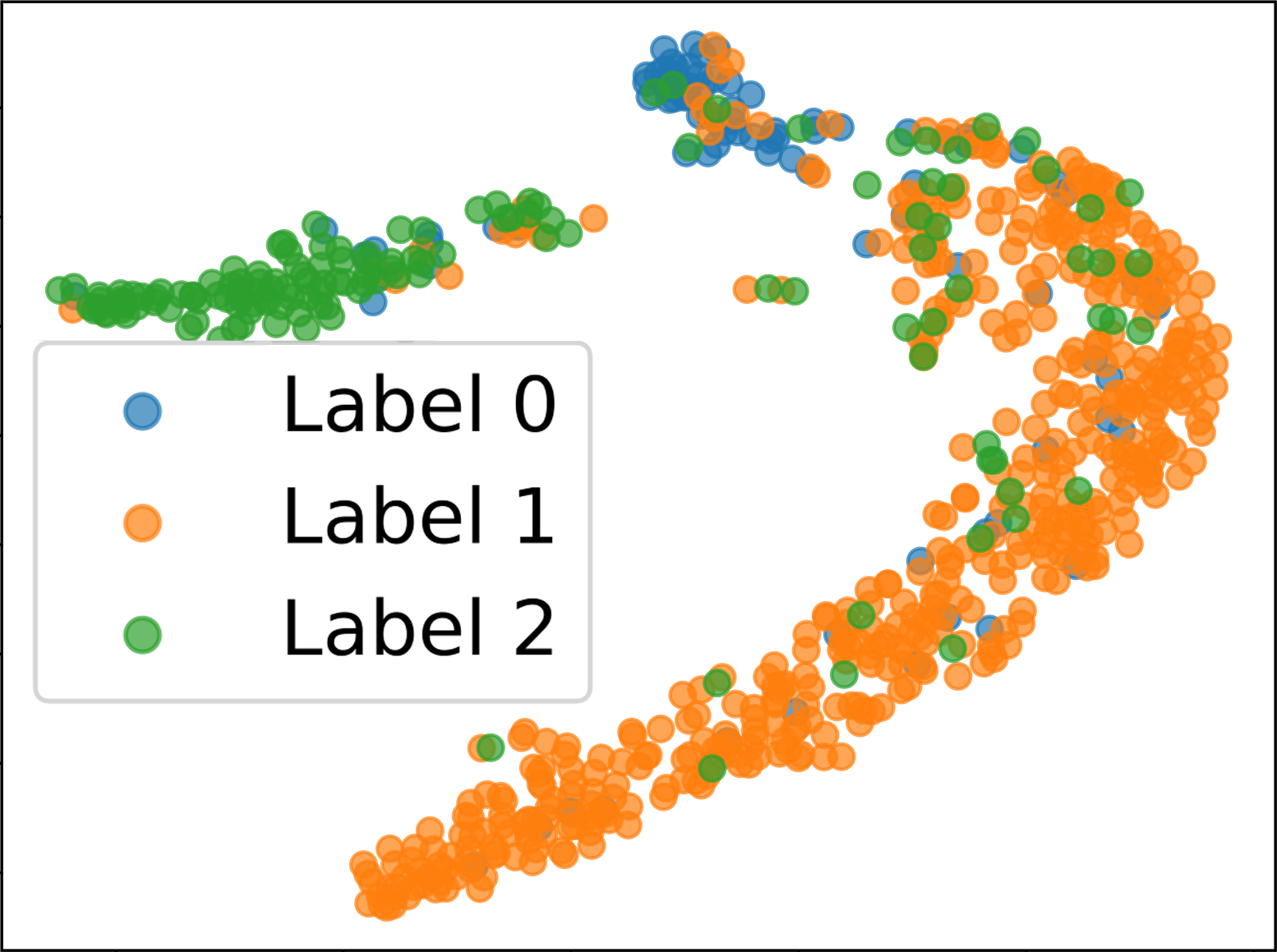}
      \caption{Task 2.}
    \end{subfigure}
    &
    \begin{subfigure}[b]{0.25\textwidth}
      \includegraphics[width=\textwidth]{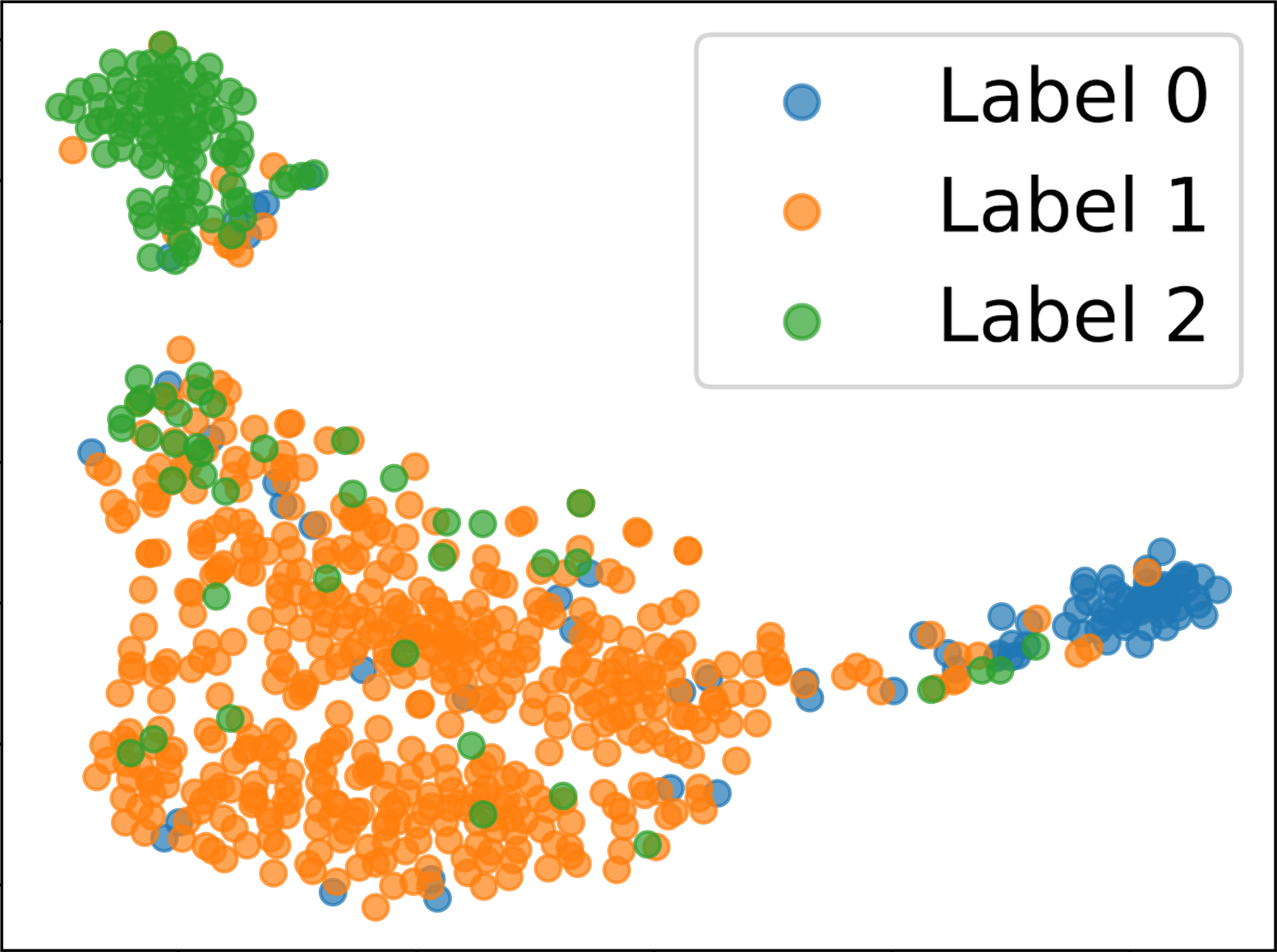}
      \caption{Task 2 (Balanced).}
    \end{subfigure}
  \end{tabular}
  
  \begin{tabular}{cc}
    \begin{subfigure}[b]{0.25\textwidth}
      \includegraphics[width=\textwidth]{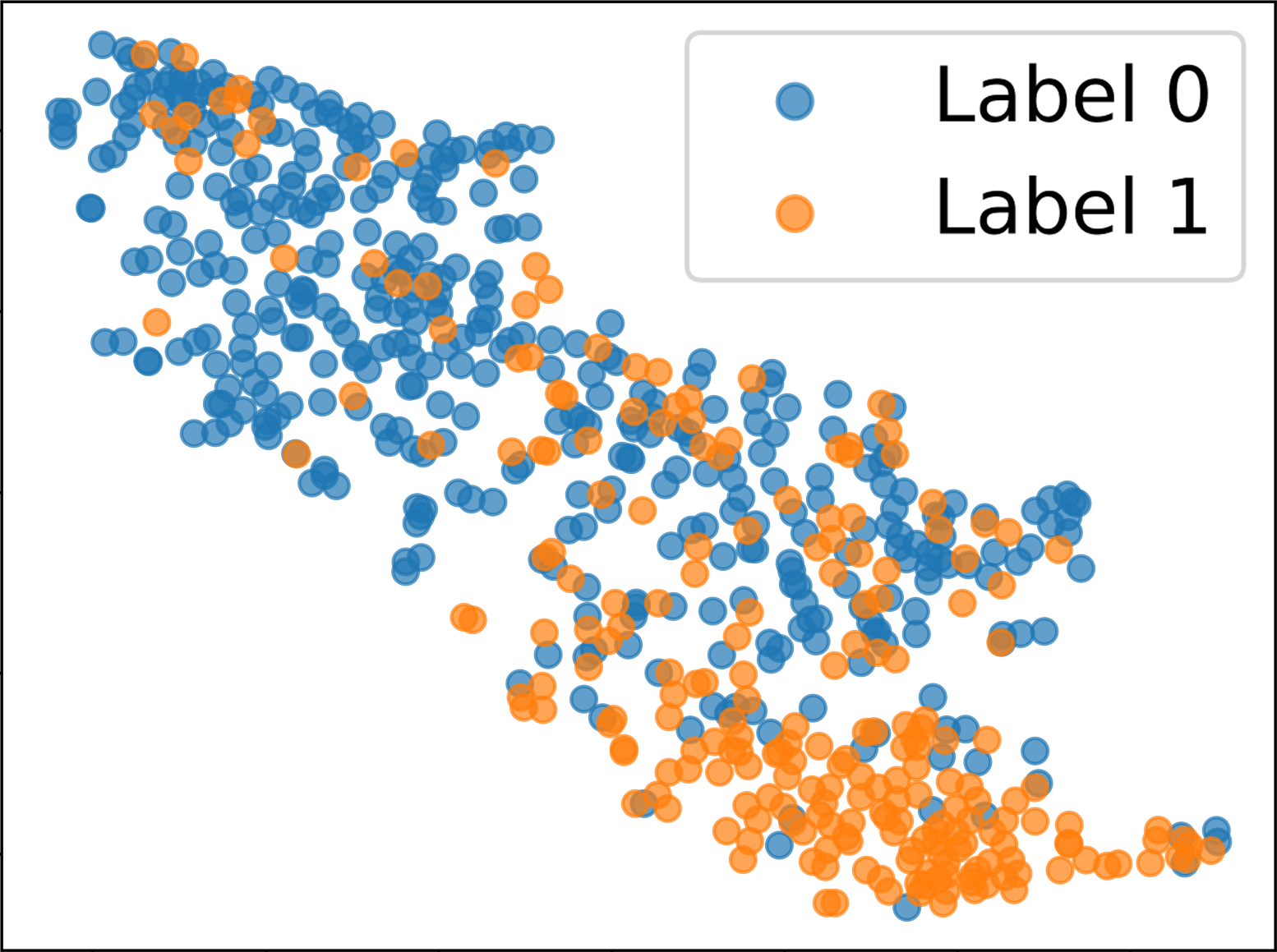}
      \caption{Task 4.}
    \end{subfigure}
    &
    \begin{subfigure}[b]{0.25\textwidth}
      \includegraphics[width=\textwidth]{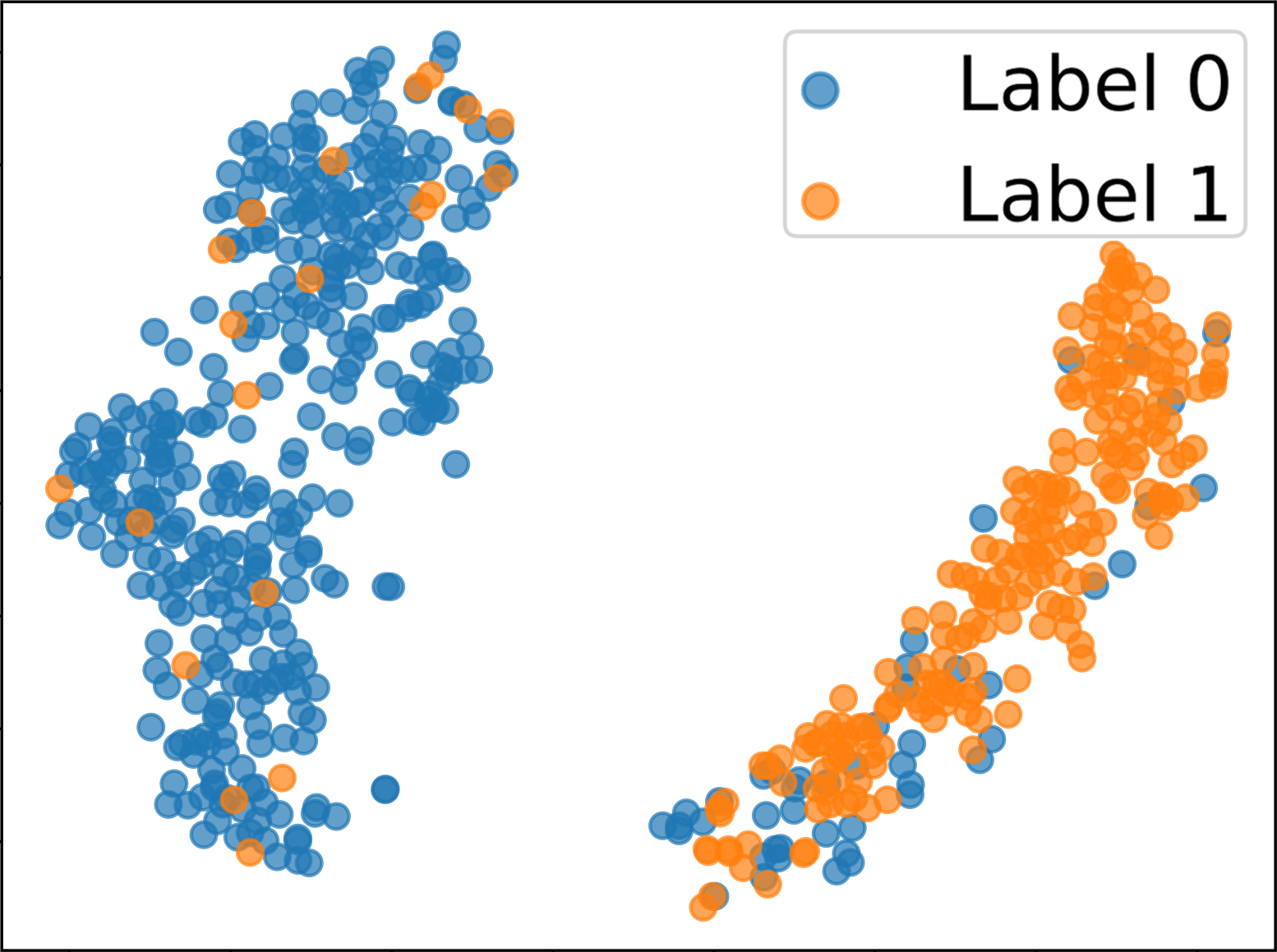}
      \caption{Task 4 (Balanced).}
    \end{subfigure}
  \end{tabular}
  \vspace{-10pt}
  \caption{Embeddings visualisations. The class separation using balanced training on the right-hand side is more clear.}
  \label{tsne}
  \vspace{-50pt}
\end{wrapfigure}

\vspace{-10pt}
\subsubsection{Visualisation of Embeddings.}\label{vis}
To validate the effectiveness of the balanced training in ALEX, the [CLS] token embeddings of the baseline method are visualised to compare with the [CLS] embeddings after the augmentation with weighted-loss fine-tuning ``(Balanced)" using t-SNE\cite{tsne}. 

As shown in Figure \ref{tsne}, it is obvious that for Task 2 and Task 4, the balanced training can successfully separate different labels' texts in different colours as the plot on the right shows. However, for Task 1, although the balanced training can also separate classes well, the effectiveness is not significant as the baseline model can also get high accuracy. Therefore, for text classification tasks that suffer from low accuracy or have low inter-class distance between each class, the balanced training method in ALEX can be proven to be effective for such public health-related problems.

\vspace{15pt}
\subsection{Effectiveness of LLMs Explanation and Correction}
In this section, the effectiveness and limitations of using LLMs for explanation and correction are analysed. Firstly, the effect of different prompts is examined by comparing a relevant prompt with an irrelevant prompt. Secondly, the effect of two strategies for the LLMs hallucinations phenomena~\cite{Ji_2023} is evaluated.

\vspace{-10pt}
\subsubsection{Case Study for Prompts.}\label{prompt}

\begin{wraptable}{r}{0.5\textwidth}
\centering \vspace{-25pt}
\caption{Case study for prompts.}\label{tab8}
{
\resizebox{1\linewidth}{!}{\begin{tabular}{c|c|c|c|c}
\toprule
Task &  \multicolumn{2}{c|}{Different Prompts} & F1 & Acc\\
\midrule
\multirow{ 6}{*}{1} & \multirow{ 3}{*}{Irr} & \multirow{ 3}{*}{\begin{tabular}[c]{@{}c@{}} ...regression results...\\...for COVID posts...\\...give reasons...\end{tabular}} & \multirow{ 3}{*}{53.39} & \multirow{ 3}{*}{59.71}\\
& & & &  \\
& & & &  \\
\cline{2-5}
& \multirow{ 3}{*}{Rel} & \multirow{ 3}{*}{\begin{tabular}[c]{@{}c@{}} ...classification...\\...for COVID posts...\\...step by step...\end{tabular}} & \multirow{ 3}{*}{\textbf{93.24}} & \multirow{ 3}{*}{\textbf{95.11}} \\
& & & &  \\
& & & &  \\
\midrule
\multirow{ 6}{*}{2} & \multirow{ 3}{*}{Irr} & \multirow{ 3}{*}{\begin{tabular}[c]{@{}c@{}} ...regression results...\\...for sentiment...\\...give reasons...\end{tabular}} & \multirow{ 3}{*}{61.93} & \multirow{ 3}{*}{47.58}\\
& & & &  \\
& & & &  \\
\cline{2-5}
& \multirow{ 3}{*}{Rel} & \multirow{ 3}{*}{\begin{tabular}[c]{@{}c@{}} ...classification...\\...for sentiment...\\...step by step...\end{tabular}} & \multirow{ 3}{*}{\textbf{89.13}} & \multirow{ 3}{*}{\textbf{77.84}} \\
& & & &  \\
& & & &  \\
\midrule
\multirow{ 6}{*}{4} & \multirow{ 3}{*}{Irr} & \multirow{ 3}{*}{\begin{tabular}[c]{@{}c@{}} ...regression results...\\...for social anxiety...\\...give reasons...\end{tabular}} & \multirow{ 3}{*}{55.34} & \multirow{ 3}{*}{61.77}\\
& & & &  \\
& & & &  \\
\cline{2-5}
& \multirow{ 3}{*}{Rel} & \multirow{ 3}{*}{\begin{tabular}[c]{@{}c@{}} ...classification...\\...for social anxiety...\\...step by step...\end{tabular}} & \multirow{ 3}{*}{\textbf{88.17}} & \multirow{ 3}{*}{\textbf{89.84}} \\
& & & &  \\
& & & &  \\
\midrule
\end{tabular}
}}
\vspace{-25pt}
\end{wraptable}
The prompt design has a significant effect on LLMs. A relevant prompt ``Rel'' can guide LLMs to better comprehend the task thus enhancing experimental performance. Conversely, an irrelevant prompt ``Irr'' may mislead LLMs which may result in erroneous corrections. In this experiment, the effect of two prompt designs is compared. The relevant prompt guides LLMs to correct the classification results, while the irrelevant prompt misguides LLMs to interpret the task as a regression task.

In Table \ref{tab8}, the results indicate a significant decrease in performance when using the irrelevant prompt which is even worse than the baseline method. Conversely, when using an appropriate prompt, the F1 score and accuracy improve by around two percent in Task 1 and Task 4. It is worth noting that even if using a relevant prompt in Task 1, the performance will still decrease compared to the previous results. Such phenomenon will be further investigated in Section \ref{hallucination}.

\vspace{-10pt}
\subsubsection{A Study for LLMs Hallucinations.}\label{hallucination}
\begin{wrapfigure}{r}{0.3\textwidth}
  \centering \vspace{-20pt}
  \includegraphics[width=0.3\textwidth]{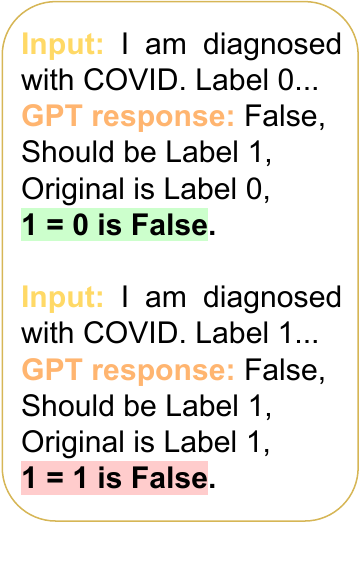}
  \vspace{-25pt}
  \caption{LLMs Hallucinations examples. The green colour denotes correct responses while the red colour denotes incorrect responses. It is obvious that the LLM generates wrong explanations for the second input.}
  \label{example}
  \vspace{-15pt}
\end{wrapfigure}

Previous experiments have indicated that GPT-3.5 exhibits balderdash when determining the correctness of labels, which may be owed to the LLMs hallucinations~\cite{Ji_2023} which refers to a phenomenon that LLMs generate responses that are actually incorrect or nonsensical. Experiments show that, in Task 1, even if GPT-3.5 successfully finds the context related to COVID and the given label is correct, it may still produce balderdash like ``1 = 1 is False'' thus return a false assertion as Figure \ref{example} shows.

Therefore, an investigation is conducted for potential causes behind LLM hallucinations. By observing those ``False'' assertion cases, it is evidently possible to find that the cause may be due to the repetitive presence of information such as ``COVID'' in the previous contexts. Similarly, researchers also found that LLMs may develop biases in their knowledge retention from the previous input, thus leading to Hallucinations~\cite{lee2022deduplicating}. For example, when the preceding input indicates the presence of COVID and is labelled as 0, the GPT-3.5 is more likely to carry this knowledge over to the subsequent input disregarding its real label. A solution is to constrain the input within a certain label to mitigate the impact of this problem as discussed in Section \ref{strategies}. 

\vspace{-25pt}
\subsubsection{Strategies for Preventing LLMs Hallucinations.} \label{strategies}
As shown in Table \ref{tab7}, directly incorporating all BERT predictions to GPT-3.5 will lead to a decreased performance. In Task 1 and Task 4, GPT-3.5 is observed to tend to classify social media posts as positive class regardless of the facts. Therefore, it may misjudge the true negative class that is predicted from BERT thus leading to decreased performance.

A possible strategy to mitigate the LLM hallucination impact is constraining it to only verify the correctness of Label 1 predictions in Task 1 and Task 4 as well as the Label 0 predictions in Task 2. Results show that in this way, GPT-3.5 can reliably enhance the F1 score of Label 1 in Task 4 as well as the accuracy in Task 2. However, for Task 1, as the balanced training Label 1's F1 score has already reached a high value of about 95, it is challenging for GPT-3.5 to discover more true positive cases. There are also many other strategies for solving LLM hallucinations like using advanced LLMs such as GPT-4 and inventing more powerful prompts to instruct LLMs for better task comprehension and language generation.

\begin{table}[!t]
\centering
\caption{Strategies for LLMs hallucinations.}\label{tab7}
\resizebox{0.6\linewidth}{!}{
\begin{tabular}{c|cc|cc|cc}
\toprule
\multirow{2}{*}{Correction} & \multicolumn{2}{c|}{Task 1} & \multicolumn{2}{c|}{Task 2}& \multicolumn{2}{c}{Task 4}\\
\cline{2-7}
& F1 & Acc & F1 & Acc & F1 & Acc \\
\midrule
All & 87.76 & 89.33 & 75.89 & 71.94 & 79.11 & 81.32\\
\midrule
Constrained & \textbf{93.24} & \textbf{95.11} & \textbf{89.13} & \textbf{77.84} & \textbf{88.17} & \textbf{89.84}\\
\bottomrule
\end{tabular}
}
\vspace{-15pt}
\end{table}

\section{Conclusion}
\vspace{-5pt}
In this paper, we focus on the promising area of analysing social media datasets for public health. To solve the imbalance problem existing in current methods, a framework involving balanced training is introduced. Moreover, the LLMs explanation and correction method is proposed to solve the limited token size problems and improve the performance of BERT. The remarkable competition results from the SMM4H 2023 competition have proven the effectiveness of our ALEX method on social media analysis for public health.

\vspace{-10pt}
\subsubsection{Acknowledgements}
The work is supported by Developing a proof-of-concept self-contact tracing app to support epidemiological investigations and outbreak response (Australia-Korea Joint Research Projects - ATSE Tech Bridge Grant).

\bibliographystyle{splncs04}
\bibliography{mybibliography}

\begin{thebibliography}{10}
\providecommand{\url}[1]{\texttt{#1}}
\providecommand{\urlprefix}{URL }
\providecommand{\doi}[1]{https://doi.org/#1}

\bibitem{data1}
A, K., S, K., K, O., G, G.H.: Twitter: An annotated data set, deep neural network classifiers, and a large-scale cohort. JOURNAL OF MEDICAL INTERNET RESEARCH  (2023)

\bibitem{COVID}
Al-Dmour, H., Masa’deh, R., Salman, A., Abuhashesh, M., Al-Dmour, R.: Influence of social media platforms on public health protection against the covid-19 pandemic via the mediating effects of public health awareness and behavioral changes: integrated model. Journal of medical Internet research  \textbf{22}(8),  e19996 (2020)

\bibitem{al2022role}
Al-Garadi, M.A., Yang, Y.C., Sarker, A.: The role of natural language processing during the covid-19 pandemic: Health applications, opportunities, and challenges. In: Healthcare. MDPI (2022)

\bibitem{inproceedings}
Bacelar{-}Nicolau, L.: The still untapped potential of social media for health promotion: The {WHO} example. In: PDH (2019)

\bibitem{brown2020language}
Brown, T.B., Mann, B., Ryder, N., Subbiah, M., Kaplan, J., Dhariwal, P., Neelakantan, A., Shyam, P., Sastry, G., Askell, A., Agarwal, S., Herbert{-}Voss, A., Krueger, G., Henighan, T., Child, R., Ramesh, A., Ziegler, D.M., Wu, J., Winter, C., Hesse, C., Chen, M., Sigler, E., Litwin, M., Gray, S., Chess, B., Clark, J., Berner, C., McCandlish, S., Radford, A., Sutskever, I., Amodei, D.: Language models are few-shot learners. CoRR  \textbf{abs/2005.14165} (2020)

\bibitem{bert}
Devlin, J., Chang, M., Lee, K., Toutanova, K.: {BERT:} pre-training of deep bidirectional transformers for language understanding. CoRR  \textbf{abs/1810.04805} (2018)

\bibitem{devlin2019bert}
Devlin, J., Chang, M., Lee, K., Toutanova, K.: {BERT:} pre-training of deep bidirectional transformers for language understanding. In: NAACL-HLT (2019)

\bibitem{9515273}
Ge, H., Zheng, S., Wang, Q.: Based bert-bilstm-att model of commodity commentary on the emotional tendency analysis. In: BDAI (2021)

\bibitem{loss}
Ho, Y., Wookey, S.: The real-world-weight cross-entropy loss function: Modeling the costs of mislabeling. {IEEE} Access  \textbf{8},  4806--4813 (2020)

\bibitem{chinchilla}
Hoffmann, J., Borgeaud, S., Mensch, A., Buchatskaya, E., Cai, T., Rutherford, E., de~Las~Casas, D., Hendricks, L.A., Welbl, J., Clark, A., Hennigan, T., Noland, E., Millican, K., van~den Driessche, G., Damoc, B., Guy, A., Osindero, S., Simonyan, K., Elsen, E., Rae, J.W., Vinyals, O., Sifre, L.: Training compute-optimal large language models. CoRR  \textbf{abs/2203.15556} (2022)

\bibitem{Ji_2023}
Ji, Z., Lee, N., Frieske, R., Yu, T., Su, D., Xu, Y., Ishii, E., Bang, Y., Madotto, A., Fung, P.: Survey of hallucination in natural language generation. ACM  \textbf{55},  248:1--248:38 (2023)

\bibitem{jiang2023uq}
Jiang, Y., Qiu, R., Zhang, Y., Huang, Z.: Uq at \#smm4h 2023: Alex for public health analysis with social media. In: AMIA (2023)

\bibitem{bertRCNN}
Kaur, K., Kaur, P.: Bert-rcnn: an automatic classification of app reviews using transfer learning based rcnn deep model (2023)

\bibitem{kumar2021classification}
Kumar, P., Bhatnagar, R., Gaur, K., Bhatnagar, A.: Classification of imbalanced data: review of methods and applications. In: IOP (2021)

\bibitem{lan2020albert}
Lan, Z., Chen, M., Goodman, S., Gimpel, K., Sharma, P., Soricut, R.: {ALBERT:} {A} lite {BERT} for self-supervised learning of language representations. In: ICLR (2020)

\bibitem{DBLP:conf/lrec/LeVFSCLACBS20}
Le, H., Vial, L., Frej, J., Segonne, V., Coavoux, M., Lecouteux, B., Allauzen, A., Crabb{\'{e}}, B., Besacier, L., Schwab, D.: Flaubert: Unsupervised language model pre-training for french. In: LREC (2020)

\bibitem{lee2022deduplicating}
Lee, K., Ippolito, D., Nystrom, A., Zhang, C., Eck, D., Callison{-}Burch, C., Carlini, N.: Deduplicating training data makes language models better. In: ACL (2022)

\bibitem{liu2005text}
Liu, N., Zhang, B., Yan, J., Chen, Z., Liu, W., Bai, F., Chien, L.: Text representation: From vector to tensor. In: ICDM (2005)

\bibitem{liu2016recurrent}
Liu, P., Qiu, X., Huang, X.: Recurrent neural network for text classification with multi-task learning. In: IJCAI (2016)

\bibitem{liu2019roberta}
Liu, Y., Ott, M., Goyal, N., Du, J., Joshi, M., Chen, D., Levy, O., Lewis, M., Zettlemoyer, L., Stoyanov, V.: Roberta: {A} robustly optimized {BERT} pretraining approach. CoRR  \textbf{abs/1907.11692} (2019)

\bibitem{loshchilov2019decoupled}
Loshchilov, I., Hutter, F.: Decoupled weight decay regularization. In: ICLR (2019)

\bibitem{tsne}
van~der Maaten, L., Hinton, G.: Viualizing data using t-sne. Journal of Machine Learning Research  \textbf{9},  2579--2605 (2008)

\bibitem{DBLP:conf/acl/MartinMSDRCSS20}
Martin, L., M{\"{u}}ller, B., Su{\'{a}}rez, P.J.O., Dupont, Y., Romary, L., de~la Clergerie, {\'{E}}., Seddah, D., Sagot, B.: Camembert: a tasty french language model. In: ACL (2020)

\bibitem{morris2020textattack}
Morris, J.X., Lifland, E., Yoo, J.Y., Grigsby, J., Jin, D., Qi, Y.: Textattack: {A} framework for adversarial attacks, data augmentation, and adversarial training in {NLP}. In: EMNLP (2020)

\bibitem{müller2020covidtwitterbert}
M{\"{u}}ller, M., Salath{\'{e}}, M., Kummervold, P.E.: Covid-twitter-bert: {A} natural language processing model to analyse {COVID-19} content on twitter. CoRR  \textbf{abs/2005.07503} (2020)

\bibitem{nguyen2020bertweet}
Nguyen, D.Q., Vu, T., Nguyen, A.T.: Bertweet: {A} pre-trained language model for english tweets. In: EMNLP (2020)

\bibitem{openai2023gpt4}
OpenAI: {GPT-4} technical report. CoRR  \textbf{abs/2303.08774} (2023)

\bibitem{instructgpt}
Ouyang, L., Wu, J., Jiang, X., Almeida, D., Wainwright, C.L., Mishkin, P., Zhang, C., Agarwal, S., Slama, K., Ray, A., Schulman, J., Hilton, J., Kelton, F., Miller, L., Simens, M., Askell, A., Welinder, P., Christiano, P.F., Leike, J., Lowe, R.: Training language models to follow instructions with human feedback. CoRR  \textbf{abs/2203.02155} (2022)

\bibitem{qiu2020pre}
Qiu, X., Sun, T., Xu, Y., Shao, Y., Dai, N., Huang, X.: Pre-trained models for natural language processing: {A} survey. CoRR  \textbf{abs/2003.08271} (2020)

\bibitem{radford2018improving}
Radford, A., Narasimhan, K., Salimans, T., Sutskever, I., et~al.: Improving language understanding by generative pre-training. OpenAI  (2018)

\bibitem{radford_language_2019}
Radford, A., Wu, J., Child, R., Luan, D., Amodei, D., Sutskever, I., et~al.: Language models are unsupervised multitask learners. OpenAI blog  \textbf{1}(8), ~9 (2019)

\bibitem{safaya2020kuisail}
Safaya, A., Abdullatif, M., Yuret, D.: {KUISAIL} at semeval-2020 task 12: {BERT-CNN} for offensive speech identification in social media. In: SemEval (2020)

\bibitem{sanh2020distilbert}
Sanh, V., Debut, L., Chaumond, J., Wolf, T.: Distilbert, a distilled version of {BERT:} smaller, faster, cheaper and lighter. CoRR  \textbf{abs/1910.01108} (2019)

\bibitem{seo2018neural}
Seo, M.J., Min, S., Farhadi, A., Hajishirzi, H.: Neural speed reading via skim-rnn. In: ICLR (2018)

\bibitem{singhal2022large}
Singhal, K., Azizi, S., Tu, T., Mahdavi, S.S., Wei, J., Chung, H.W., Scales, N., Tanwani, A.K., Cole{-}Lewis, H., Pfohl, S., Payne, P., Seneviratne, M., Gamble, P., Kelly, C., Sch{\"{a}}rli, N., Chowdhery, A., Mansfield, P.A., y~Arcas, B.A., Webster, D.R., Corrado, G.S., Matias, Y., Chou, K., Gottweis, J., Tomasev, N., Liu, Y., Rajkomar, A., Barral, J.K., Semturs, C., Karthikesalingam, A., Natarajan, V.: Large language models encode clinical knowledge. CoRR  \textbf{abs/2212.13138} (2022)

\bibitem{sun2020finetune}
Sun, C., Qiu, X., Xu, Y., Huang, X.: How to fine-tune {BERT} for text classification? In: CCL (2019)

\bibitem{llama}
Touvron, H., Lavril, T., Izacard, G., Martinet, X., Lachaux, M., Lacroix, T., Rozi{\`{e}}re, B., Goyal, N., Hambro, E., Azhar, F., Rodriguez, A., Joulin, A., Grave, E., Lample, G.: Llama: Open and efficient foundation language models. CoRR  \textbf{abs/2302.13971} (2023)

\bibitem{vaswani2017attention}
Vaswani, A., Shazeer, N., Parmar, N., Uszkoreit, J., Jones, L., Gomez, A.N., Kaiser, L., Polosukhin, I.: Attention is all you need. CoRR  \textbf{abs/1706.03762} (2017)

\bibitem{chain}
Wei, J., Wang, X., Schuurmans, D., Bosma, M., Ichter, B., Xia, F., Chi, E.H., Le, Q.V., Zhou, D.: Chain-of-thought prompting elicits reasoning in large language models. In: NeurIPS (2022)

\bibitem{wolf2020huggingfaces}
Wolf, T., Debut, L., Sanh, V., Chaumond, J., Delangue, C., Moi, A., Cistac, P., Rault, T., Louf, R., Funtowicz, M., Brew, J.: Huggingface's transformers: State-of-the-art natural language processing. CoRR  \textbf{abs/1910.03771} (2019)

\bibitem{u2l}
Wu, D., Zhang, B., Yang, C., Peng, Z., Xia, W., Chen, X., Lei, X.: {U2++:} unified two-pass bidirectional end-to-end model for speech recognition. CoRR  \textbf{abs/2106.05642} (2021)

\bibitem{DBLP:journals/corr/abs-1906-08237}
Yang, Z., Dai, Z., Yang, Y., Carbonell, J.G., Salakhutdinov, R., Le, Q.V.: Xlnet: Generalized autoregressive pretraining for language understanding. CoRR  \textbf{abs/1906.08237} (2019)

\bibitem{yogatama2017generative}
Yogatama, D., Dyer, C., Ling, W., Blunsom, P.: Generative and discriminative text classification with recurrent neural networks. CoRR  \textbf{abs/1703.01898} (2017)

\bibitem{glm}
Zeng, A., Liu, X., Du, Z., Wang, Z., Lai, H., Ding, M., Yang, Z., Xu, Y., Zheng, W., Xia, X., Tam, W.L., Ma, Z., Xue, Y., Zhai, J., Chen, W., Zhang, P., Dong, Y., Tang, J.: {GLM-130B:} an open bilingual pre-trained model. CoRR  \textbf{abs/2210.02414} (2022)

\bibitem{ZENG2021437}
Zeng, D., Cao, Z., Neill, D.B.: Chapter 22 - artificial intelligence–enabled public health surveillance—from local detection to global epidemic monitoring and control. In: AIM (2021)

\bibitem{zhang2022twhinbert}
Zhang, X., Malkov, Y., Florez, O., Park, S., McWilliams, B., Han, J., El{-}Kishky, A.: Twhin-bert: {A} socially-enriched pre-trained language model for multilingual tweet representations. CoRR  \textbf{abs/2209.07562} (2022)

\bibitem{DBLP:journals/corr/abs-2303-18223}
Zhao, W.X., Zhou, K., Li, J., Tang, T., Wang, X., Hou, Y., Min, Y., Zhang, B., Zhang, J., Dong, Z., Du, Y., Yang, C., Chen, Y., Chen, Z., Jiang, J., Ren, R., Li, Y., Tang, X., Liu, Z., Liu, P., Nie, J., Wen, J.: A survey of large language models. CoRR  \textbf{abs/2303.18223} (2023)

\end{thebibliography}

\end{document}